\documentclass[11pt,a4paper]{article}
\usepackage[hyperref]{emnlp-ijcnlp-2019}
\usepackage{times}
\usepackage{latexsym}
\usepackage{graphicx}
\usepackage{multirow}
\usepackage{url}
\usepackage{amssymb}
\usepackage{amsmath}
\usepackage{enumitem}
\usepackage{url}
\usepackage[normalem]{ulem}
\useunder{\uline}{\ul}{}

\aclfinalcopy 

\title{MoEL: Mixture of Empathetic Listeners}
\author{Zhaojiang Lin, Andrea Madotto, Jamin Shin, Peng Xu, Pascale Fung \\
Center for Artificial Intelligence Research (CAiRE)\\
  Department of Electronic and Computer Engineering\\
  The Hong Kong University of Science and Technology, Clear Water Bay, Hong Kong\\
  \texttt{\{zlinao,amadotto,jmshinaa,pxuab\}@connect.ust.hk},\\ \texttt{ pascale@ece.ust.hk}}

\date{}

\begin{document}
\maketitle
\begin{abstract}
Previous research on empathetic dialogue systems has mostly focused on generating responses given certain emotions. However, being empathetic not only requires the ability of generating emotional responses, but more importantly, requires the understanding of user emotions and replying appropriately. In this paper, we propose a novel end-to-end approach for modeling empathy in dialogue systems: Mixture of Empathetic Listeners (MoEL). Our model first captures the user emotions and outputs an emotion distribution. Based on this, MoEL will \textit{softly combine} the output states of the \textit{appropriate} Listener(s), which are each optimized to react to certain emotions, and generate an empathetic response.
Human evaluations on \textit{empathetic-dialogues}~\cite{rashkin2018know} dataset confirm that MoEL outperforms multitask training baseline in terms of empathy, relevance, and fluency. Furthermore, the case study on generated responses of different Listeners shows high interpretability of our model.
\end{abstract}

\section{Introduction}

Neural network approaches for conversation models have shown to be successful in scalable training and generating fluent and relevant responses~\cite{vinyals2015neural}. However, it has been pointed out by~\citet{li2016diversity, li2016persona, li2016deep, wu2018response} that only using Maximum Likelihood Estimation as the objective function tends to lead to \textit{generic} and \textit{repetitive} responses like ``I am sorry''. Furthermore, many others have shown that the incorporation of additional inductive bias leads to a more engaging chatbot, such as understanding commonsense~\cite{dinan2018wizard}, or modeling consistent persona~\cite{li2016persona, zhang2018personalizing, mazare2018training}. 

\begin{table}[ht]
\begin{tabular}{rl}
\hline
\multicolumn{2}{c}{\textbf{Emotion: Angry}} \\ \hline \hline
\multicolumn{2}{c}{\begin{tabular}[c]{@{}c@{}}\textbf{Situation}\\ I was furious when I got in \\ my first car wreck.\end{tabular}} \\ \hline \hline
\multicolumn{1}{r|}{\textbf{Speaker}} & \begin{tabular}[c]{@{}l@{}}I was driving on the interstate and \\ another car ran into the back of me.\end{tabular} \\ \hline
\multicolumn{1}{r|}{\textbf{Listener}} & \begin{tabular}[c]{@{}l@{}}Wow. Did you get hurt? \\ Sounds scary.\end{tabular} \\ \hline
\multicolumn{1}{r|}{\textbf{Speaker}} & \begin{tabular}[c]{@{}l@{}}No just the airbags went off and\\ I hit my head and got a few bruises.\end{tabular} \\ \hline
\multicolumn{1}{r|}{\textbf{Listener}} & \begin{tabular}[c]{@{}l@{}}I am always scared about those\\ airbags! I am so glad you are ok!\end{tabular} \\ \hline
\end{tabular}
\caption{One conversation from empathetic dialogue, a speaker tells the situation he(she) is facing, and a listener try to understand speaker's feeling and respond accordingly}
\label{tab:example_data}
\end{table}

Meanwhile, another important aspect of an engaging human conversation that received relatively less focus is emotional understanding and empathy~\cite{rashkin2018know, dinan2019second, wolf2019transfertransfo}. Intuitively, ordinary social conversations between two humans are often about their daily lives that revolve around happy or sad experiences. In such scenarios, people generally tend to respond in a way that acknowledges the feelings of their conversational partners.

Table~\ref{tab:example_data} shows an conversation from the \textit{empathetic-dialogues} dataset~\cite{rashkin2018know} about how an empathetic person would respond to the stressful situation the \textit{Speaker} has been through.
However, despite the importance of empathy and emotional understanding in human conversations, it is still very challenging to train a dialogue agent able to recognize and respond with the correct emotion.
\begin{figure*}[t]
\centering
\includegraphics[width=0.80\linewidth]{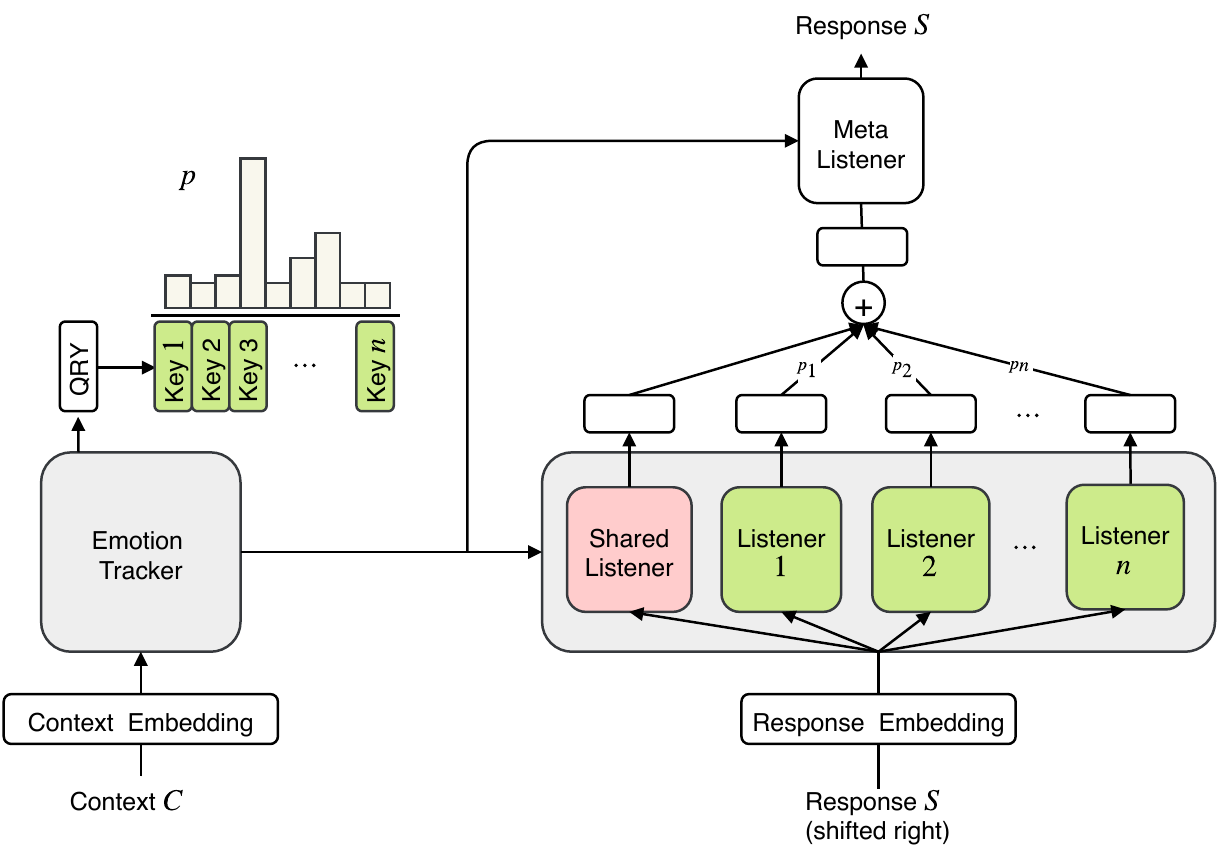}
\caption{The proposed model Mixture of Empathetic Listeners, which has an emotion tracker, $n$ empathetic listeners along with a shared listener, and a meta listener to fuse the information from listeners and produce the empathetic response.}
\label{fig:moel}
\end{figure*}

So far, to solve the problem of empathetic dialogue response generation, which is to understand the user emotion and respond appropriately~\cite{bertero2016real}, there have been mainly two lines of work. The first is a multi-task approach that jointly trains a model to predict the current emotional state of the user and generate an appropriate response based on the state~\cite{lubis2018eliciting,rashkin2018know}. Instead, the second line of work focuses on conditioning the response generation to a certain fixed emotion~\cite{hu2017toward,wang2018sentigan,zhou2018mojitalk,zhou2018emotional}.

Both cases have succeeded in generating empathetic and emotional responses, but have neglected some crucial points in empathetic dialogue response generation. 1) The first assumes that by understanding the emotion, the model implicitly learns how to respond appropriately. However, without any additional inductive bias, a single decoder learning to respond for all emotions will not only lose interpretability in the generation process, but will also promote more generic responses. 2) The second assumes that the emotion to condition the generation on is given as input, but we often do not know which emotion is appropriate in order to generate an empathetic response.

Therefore, in this paper, to address the above issues, we propose a novel end-to-end empathetic dialogue agent, called Mixture of Empathetic Listeners~\footnote{The code will be released at \url{https://github.com/HLTCHKUST/MoEL}} (MoEL) inspired by~\citet{shazeer2017outrageously}. Similar to \citet{rashkin2018know}, we first encode the dialogue context and use it to recognize the emotional state ($n$ possible emotions). However, the main difference is that our model consists of $n$ decoders, further denoted as \textit{listeners}, which are optimized to react to each context emotion accordingly. The listeners are trained along with a Meta-listener that \textit{softly combines} the output decoder states of each listener according to the emotion classification distribution. Such design allows our model to explicitly learn how to choose an appropriate reaction based on its understanding of the context emotion. A detailed illustration of MoEL is shown in Figure~\ref{fig:moel}.

The proposed model is tested against several competitive baseline settings~\cite{vaswani2017attention,rashkin2018know}, and evaluated with human judges. The experimental results show that our approach outperforms the baselines in both empathy and relevance. Finally, our analysis demonstrates that not only MoEL effectively attends to the right listener, but also each listener learns how to properly react to its corresponding emotion, hence allowing a more interpretable generative process.

\section{Related Work}
\paragraph{Conversational Models:} Open domain conversational models has been widely studied~\cite{serban2016generative,vinyals2015neural,wolf2019transfertransfo}. A recent trend is to produce personalized responses by conditioning the generation on a persona profile to make the response more consistent through the dialogue~\cite{li2016persona}. In particular, PersonaChat~\cite{personachat,kulikov2018importance} dataset was created, and then extended in ConvAI 2 challenge~\cite{dinan2019second}, to show that by adding persona information as input to the model, the produced responses elicit more consistent personas. 
Based on such, several follow-up work has been presented~\cite{millionspersona, hancock2019learning,  joshi2017personalization, kulikov2018importance,yavuz2018deepcopy,zemlyanskiy2018aiming,madotto2019personalizing}.
However, such personalized dialogue agents focus only on modeling a consistent persona and often neglect the feelings of their conversation partners.

Another line of work combines retrieval and generation to promote the response diversity~\cite{cai2018skeleton,weston2018retrieve,wu2018response}. However, only fewer works focus on emotion~\cite{winata2017nora,winata2019caire_hkust,xu2018emo2vec,fan2018multi,fan2018video,fan2018unsupervised,lee2019team} and empathy in the context of dialogues systems~\cite{bertero2016real,chatterjee2019understanding,SemEval2019Task3,shin2019happybot}. For generating emotional dialogues,~\citet{hu2017toward, wang2018sentigan, zhou2018mojitalk} successfully introduce a framework of controlling the sentiment and emotion of the generated response, while~\cite{zhou2018mojitalk} also introduces a new Twitter conversation dataset and propose to distantly supervised the generative model with emojis. Meanwhile,~\cite{lubis2018eliciting, rashkin2018know} also introduce new datasets for empathetic dialogues and train multi-task models on it. 

\paragraph{Mixture of Experts:} The idea of having specialized parameters, or so-called experts, has been widely studied topics in the last two decades~\cite{jacobs1991adaptive,jordan1994hierarchical}. For instance, different architectures and methodologies have been used such as SVM~\cite{collobert2002parallel}, Gaussian Processes~\cite{tresp2001mixtures,theis2015generative,deisenroth2015distributed}, Dirichlet Processes~\cite{shahbaba2009nonlinear}, Hierarchical Experts~\cite{yao2009hierarchical}, Infinite Number of Experts~\cite{rasmussen2002infinite} and sequential expert addition~\cite{aljundi2017expert}. More recently, the Mixture Of Expert~\cite{shazeer2017outrageously,kaiser2017one} model was proposed which added a large number of experts in between of two LSTM~\cite{schmidhuber:1987:srl} layers to enhance the capacity of the model. This idea of having independent specialized experts inspires our approach to model the reaction to each emotion with a separate expert.

\section{Mixture of Empathetic Listeners}
The dialogue context is an alternating set of utterances from speaker and listener. We denote the dialogue context as $C = \{U_1, S_1, U_2, S_2, \cdots, U_t\}$ and the speaker emotion state at each utterance as $Emo = \{e_1, e_2, \cdots, e_t\}$ where $\forall e_i \in \{1,\dots,n\}$. Then, our model aims to track the speaker emotional state $e_t$ from the dialogue context $C$, and generates an empathetic response $S_t$.

Overall, MoEL is composed of three components: an \textit{emotion tracker}, \textit{emotion-aware listeners}, and a \textit{meta listener} as shown in Figure~\ref{fig:moel}. The emotion tracker (which is also the context encoder) encodes $C$ and computes a distribution over the possible user emotions. Then all the listeners independently attend to this distribution to compute their own representation. Finally, the meta listener takes the weighted sum of representations from the listeners and generates the final response. 

\begin{figure*}[t]
\centering
\includegraphics[width=\linewidth]{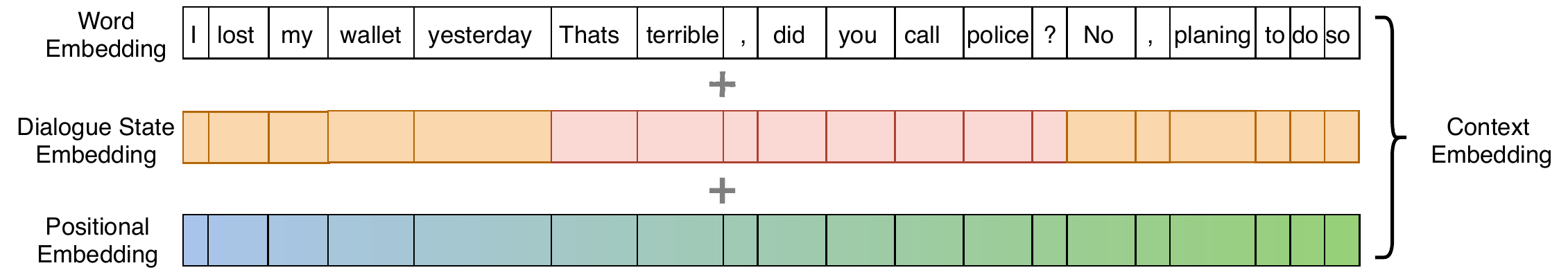}
\caption{Context embedding is computed by summing up the word embedding, dialogue state embedding and positional embedding for each token.}
\label{context_embedding}
\end{figure*}

\subsection{Embedding}
We define the context embedding $E^C\in \mathbb{R}^{|V| \times d_{emb}}$, and the response embedding $E^R\in \mathbb{R}^{|V| \times d_{emb}}$ which are used to convert tokens into embeddings. In multi-turn dialogues, ensuring that the model is able to distinguish among turns is essential, especially when multiple emotion are present in different turns. Hence, we incorporate a dialogue state embedding in the input. This is used to enable the encoder to distinguish speaker utterances and listener utterances~\cite{wolf2019transfertransfo}. As shown in Figure~\ref{context_embedding}, our context embedding $E^C$ is the positional sum of the word embedding $E^W$, the positional embedding $E^P$~\cite{vaswani2017attention} and the dialogue state embedding $E^D$.
\begin{equation}
    E^C(C) = E^W(C) + E^P(C) + E^D(C)
\end{equation}

\subsection{Emotion Tracker}
MoEL uses a standard transformer encoder~\cite{vaswani2017attention} for the emotion tracker. We first flatten all dialogue turns in $C$, and map each token into its vectorial representation using the context embedding $E^C$. Then the encoder encodes the context sequence into a context representation. 
We add a query token $QRY$ at the beginning of each input sequence as in BERT~\cite{devlin2018bert}, to compute the weighted sum of the output tensor. Denoting a transformer encoder as $TRS_{Enc}$, then corresponding context representation become:
\begin{equation}
    H = TRS_{Enc}(E^C([QRY; C]))
\end{equation}
where $[ ; ]$ denotes concatenation, $H\in \mathbb{R}^{L \times d_{model}}$ where $L$ is the sequence length. Then, we define the final representation of the token $QRY$ as
\begin{equation}
    q = H_0 
\label{q}
\end{equation}
where $q\in \mathbb{R}^{d_{model}}$, which is then used as the query for generating the emotion distribution.

\subsection{Emotion Aware Listeners}
The emotion aware listeners mainly consist of 1) a \textit{shared listener} that learns shared information for all emotions and 2) $n$ independently parameterized Transformer decoders~\cite{vaswani2017attention} that learn how to appropriately react given a particular emotional state. All the listeners are modeled by a standard transformer decoder layer block, denoted as $TRS_{Dec}$, which is made of three sub-components: a multi-head self-attention over the response input embedding, a multi-head attention over the output of the emotion tracker, and a position-wise fully connected feed-forward network. 

Thus, we define the set of listeners as $L=[TRS^0_{Dec},\dots,TRS^n_{Dec}]$. Given the target sequence shifted by one $r_{0:t-1}$, each listener compute its own emotional response representation $V_i$:
\begin{equation}
    V_i = TRS^i_{Dec}(H, E^R(r_{0:t-1})) \label{v_i}
\end{equation}
where $TRS^i_{Dec}$ refers to the $i$-th listener, including the shared one. Conceptually, we expect that the output from the shared listener, $TRS^0_{Dec}$, to be a general representation which can help the model to capture the dialogue context. On the other hand, we expect that each empathetic listener learns how to respond to a particular emotion. To model this behavior, we assign different weights to each empathetic listener according to the user emotion distribution, while assigning a fixed weight of 1 to the shared listener. 

To elaborate, we construct a Key-Value Memory Network~\cite{miller2016key} and represent each memory slot as a vector pair $(k_i,V_i)$, where $k_i\in \mathbb{R}^{d_{model}}$ denotes the key vector and $V_i$ is from Equation~\ref{v_i}. Then, the encoder informed query $q$ is used to address the key vectors $k$ by performing a dot product followed by a Softmax function. Thus, we have:

\begin{equation}
    p_{i} = \frac{e^{q^{\top}k_i}}{\sum_{j=1}^{n} e^{q^{\top}k_j}}
\label{key}
\end{equation}
each $p_{i}$ is the score assigned to $V_i$, thus used as the weight of each listener. During training, given the speaker emotion state $e_t$, we supervise each weight $p_i$ by maximizing the probability of the emotion state $e_{t}$ with a cross entropy loss function:
\begin{equation}
    \mathcal{L}_{1} = - \log p_{e_t}
\end{equation}

\noindent Finally, the combined output representation is compute by the weighted sum of the memory values $V_i$ and the shared listener output $V_0$.
\begin{equation}
    V_M = V_0 + \sum_{i=1}^n p_{i}V_i
\end{equation}
\begin{figure*}[t]
\centering
\includegraphics[width=\linewidth]{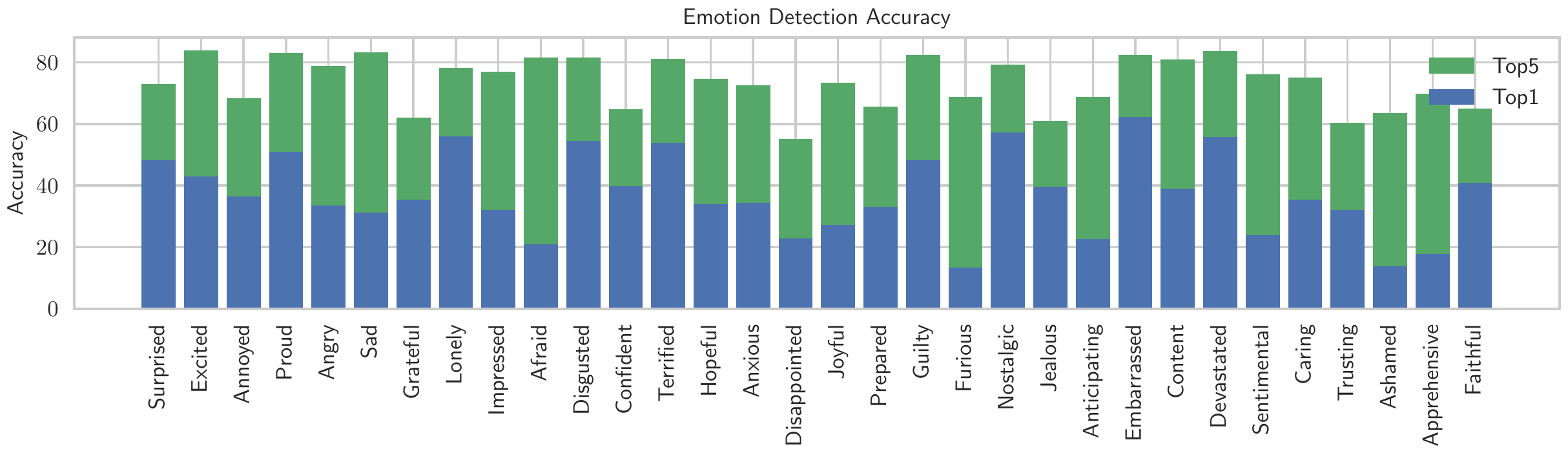}
\caption{Top-1 and Top-5 emotion detection accuracy over 32 emotions at each turn}
\label{acc}
\end{figure*}

\subsection{Meta Listener}
Finally, the Meta Listener is implemented using another transformer decoder layer, which further transform the representation of the listeners and generates the final response. The intuition is that each listener specializes to a certain emotion and the Meta Listener gathers the opinions generated by multiple listeners to produce the final response. Hence, we define another $TRS^{Meta}_{Dec}$, and an affine transformation $W\in \mathbb{R}^{d_{model} \times |V|}$ to compute: 
\begin{equation}
    O = TRS^{Meta}_{Dec}(H, V_M)
\end{equation}
\begin{equation}
    p(r_{1:t}|C,r_{0:t-1}) = \text{softmax}(O^{\top} W)
\end{equation}
where $O\in \mathbb{R}^{d_{model} \times t}$ is the output of meta listener and $p(r_{1:t}|C,r_{0:t-1})$ is a distribution over the vocabulary for the next tokens. We then use a standard maximum likelihood estimator (MLE) to optimize the response prediction:
\begin{equation}
\mathcal{L}_{2}=- \log p\left(S_{t} | C \right)
\label{mle}
\end{equation}
Lastly, all the parameters are jointly trained end-to-end to optimize the listener selection and response generation by minimizing the weighted-sum of two losses:
\begin{equation}
\mathcal{L} = \alpha\mathcal{L}_{1} + \beta\mathcal{L}_{2}
\end{equation}
Where $\alpha$ and $\beta$ are hyperparameters to balance two loss.
\begin{table}[t]
\resizebox{\linewidth}{!}{
\begin{tabular}{r|c|c|c|c|c}
\hline
                    & \textbf{Params.} & \textbf{BLEU} & \textbf{Empathy} & \textbf{Relevance} & \textbf{Fluency} \\ \hline
\textit{Gold}            & -    & - & 3.93   & 3.93    & 3.35 \\ \hline
\textit{TRS}     & 16.94M        & 3.02 & 3.32   & 3.47    & \textbf{3.52}   \\
\textit{MultiTRS}    & 16.95M   & 2.92  & 3.36 & 3.57   & 3.31       \\
\textit{MoEL}   & 23.1M      & 2.90 & \textbf{3.44}   & \textbf{3.70}     & 3.47   \\ \hline
\end{tabular}}
\caption{\label{tab:evaluation} Comparison between our proposed methods and baselines. All of models receive close BLEU score. MoEL achieve highest \textit{Empathy} and \textit{Relevance} score, while TRS achieve better \textit{Fluency} score. The number of parameters for each model is reported.} 
\end{table}
\begin{table}[t]
\begin{tabular}{r|ccc}
\hline
\multicolumn{1}{c|}{\textbf{Model}} & \textbf{Win} & \textbf{Loss} & \textbf{Tie} \\ \hline
\textit{MoEL vs TRS} & 37.3\% & 18.7\% & 44\% \\ \hline
\textit{MoEL vs Multi-TRS} & 36.7\% & 32.6\% & 30.7\% \\ \hline
\end{tabular}
\caption{Result of human A/B test. Tests are conducted pairwise between MoEL and baseline models}
\end{table}

\section{Experiment}
\subsection{Dataset}
We conduct our experiment on the \textit{empathetic-dialogues}~\cite{rashkin2018know} dataset which consist of 25k one-to-one open-domain conversation grounded in emotional situations. The dataset provides 32, evenly distributed, emotion labels. Table~\ref{tab:example_data} shows an example from the training set. The speakers are talking about their situation and the listeners is trying to understand their feeling and reply accordingly. At training time the emotional labels of the speakers are given, while we hide the label in test time to evaluate the \textit{empathy} of our model.

\subsection{Training}
We train our model using Adam optimizer~\cite{kingma2014adam} and varied the learning rate during training following~\cite{vaswani2017attention}. The weight of both losses $\alpha$ and $\beta$ are set to 1 for simplicity. We use pre-trained Glove vectors~\cite{pennington2014glove} to initialize the word embedding and we share it across the encoder and the decoder. The rest of the parameters are randomly initialized. 

In the early training stage, emotion tracker randomly assign weights to the listeners, and may send noisy gradient flow back to the wrong listeners, which can make the model convergence harder. To stabilize the learning process, we replace the distribution $p$ of the listeners with the oracle emotion $e_t$ information using a certain probability $\epsilon_{oracle}$, and we gradually anneal it during the training. We set an annealing rate $\gamma=1\times10^{-3}$, and a threshold $t_{thd}$ equal to $1\times10^{4}$, thus at each iteration $t$ iteration we compute:

\begin{equation}
\epsilon_{oracle} = \gamma + (1-\gamma) e^{-\frac{t}{t_{thd}}}
\end{equation}

\begin{table*}[t]
\begin{tabular}{p{1.8cm}|p{12cm}}
\hline
\textbf{Emotion}     & Disgusted                                                                                                                                                                    \\ \hline
\textbf{Context}   & \begin{tabular}[c]{@{}l@{}}\textbf{Speaker}:  i had to dog-sit for my dad about a week ago . i had to clean up dog \\poop at least 10 times , it was awful .\end{tabular} \\ \hline
\textbf{TRS}       & that is \textbf{horrible} . i hope you did not have to clean it up .                                                                                                                         \\ \hline
\textbf{Multi-TRS} & oh my goodness , that must have been so \textbf{scary} !                                                                                                                                     \\ \hline
\textbf{MoEL}      & oh wow , that is pretty \textbf{gross} .                                                                                                                                                     \\ \hline
\textbf{Goal}      & ohh , cleaning dog poop is always been annoying , yet petting the dog is always amusing                                                                                        \\ \hline

\hline
\hline
\textbf{Emotion}     & Disappointed                                                                                                                                                                                                                                                                   \\ \hline
\textbf{Context}   & \begin{tabular}[c]{@{}l@{}}\textbf{Speaker}: i was turned down for a job i really thought i would be a good fit at . \\ \textbf{Listener}: did they give you any reason for not giving you the job ? \\ \textbf{Speaker}: someone else interviewed who had much more experience than i did .\end{tabular} \\ \hline
\textbf{TRS}       & that is good . i hope you have a good time .                                                                                                                                                                                                                                   \\ \hline
\textbf{Multi-TRS} & that is a good idea . i hope it was a good idea .                                                                                                                                                                                                                              \\ \hline
\textbf{MoEL}      & i am sorry to hear that . i am sure you will do great again                                                                                                                                                                                                                    \\ \hline
\textbf{Goal}      & well , at least a candidate that deserved the job got it .                                                                                                                                                    \\ \hline

\hline
\noalign{\vskip 3mm}
\hline
\multicolumn{2}{l}{\begin{tabular}[c]{@{}l@{}}\textbf{Speaker}: so my friend 's baby fell down and scraped his knee yesterday and i was trying so hard\\ to cheer him up.\\ $\textbf{Listener}_{caring}$: that was really nice of you. I am glad you were able to help him.\end{tabular}} \\ \hline
\multicolumn{1}{l|}{$\textbf{Listener}_{terrified}$} & oh wow ! that must have been so scary . \\ \hline
\multicolumn{1}{l|}{$\textbf{Listener}_{sad}$} & oh no , i am so sorry to hear that . \\ \hline
\multicolumn{1}{l|}{$\textbf{Listener}_{excited}$} & that sounds like a good idea . \\ \hline
\multicolumn{1}{l|}{$\textbf{Listener}_{proud}$} & that was awesome ! did he get any good at you ? \\ \hline
\hline
\multicolumn{2}{l}{\begin{tabular}[c]{@{}l@{}}\textbf{Speaker}: my husband lost a job but I am hoping he can find a full-time job soon.\\ $\textbf{Listener}_{hopeful}$: what kind of job is it?\end{tabular}} \\ \hline
\multicolumn{1}{l|}{$\textbf{Listener}_{terrified}$} & oh that sucks . you should have been together . \\ \hline
\multicolumn{1}{l|}{$\textbf{Listener}_{sad}$} & i am so sorry . i am sure he will be fine . \\ \hline
\multicolumn{1}{l|}{$\textbf{Listener}_{excited}$} & that sounds awesome . what kind of job did he get you for ? \\ \hline
\multicolumn{1}{l|}{$\textbf{Listener}_{proud}$} & oh wow ! congratulations to him . you must be proud of him . \\ \hline
\end{tabular}
\caption{Generated responses from TRS, Multi-TRS and MoEL in 2 different user emotion states (\textbf{top}) and comparing generation from different listeners (\textbf{bottom}). We use hard attention on Terrified, Sad, Excited and Proud listeners.}
\label{compare}
\end{table*}

\subsection{Baseline}
We compare our model with two baselines:
\paragraph{Transformer~(TRS)} The standard Transformer model~\cite{vaswani2017attention} that is trained to minimize MLE loss as in Equation~\ref{mle}.
\paragraph{Multitask Transformer~(Multi-TRS)} A Multitask Transformer trained as~\cite{rashkin2018know} to incorporate additional supervised information about the emotion. The encoder of multitask transformer is the same as our emotion tracker, and the context representation $Q$, from Equation~\ref{q}, is used as input to an emotion classifier. The whole model is jointy trained by optimizing both the classification and generation loss.

\begin{figure*}[t]
\centering
\includegraphics[width=\linewidth]{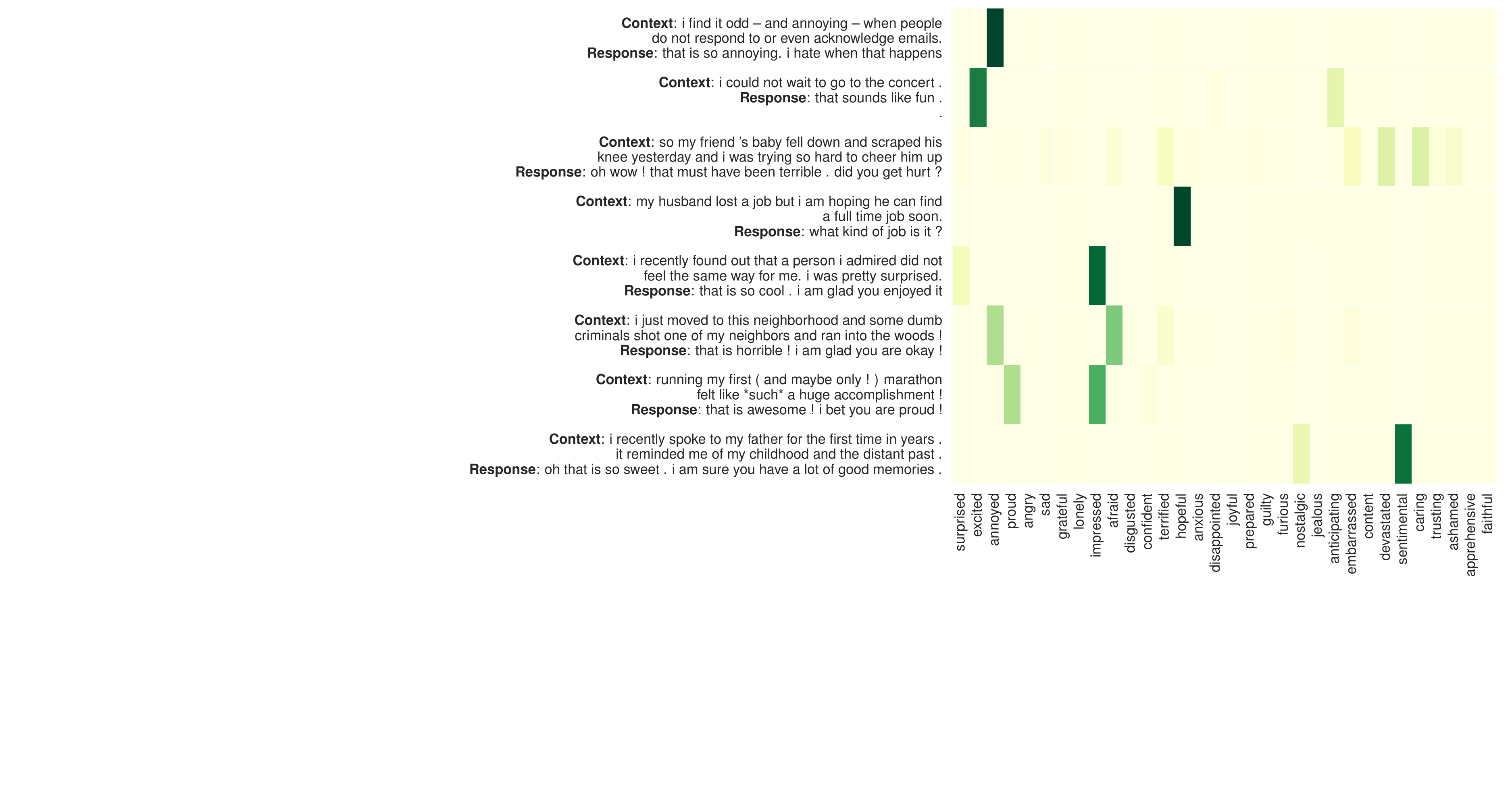}
\caption{The visualization of attention on the listeners: The left side is the context followed by the responses generated by MoEL. The heat map illustrate the attention weights on 32 listeners}
\label{viz}
\end{figure*}

\subsection{Hyperparameter}
In all of our experiments we used 300 dimensional word embedding and 300 hidden size everywhere. We use 2 self-attention layers made up of 2 attention heads each with embedding dimension 40. We replace Positionwise Feedforward sub-layer with 1D convolution with 50 filters of width 3. We train all of models with batch size 16 and we use batch size 1 in the test time.

\subsection{Evaluation Metrics}
\paragraph{BLEU} We compute BLEU scores~\cite{papineni2002bleu} to compare the generated response against human responses. However, in open-domain dialogue response generation, BLEU is not a good measurement of generation quality~\cite{liu2016not}, so we use BLEU only as a reference.

\paragraph{Human Ratings} In order to measure the quality of the generated responses, we conduct human evaluations with Amazon Mechanical Turk. Following~\citet{rashkin2018know}, we first randomly sample 100 dialogues and their corresponding generations from MoEL and the baselines. For each response, we assign three human annotators to score the following aspect of models: \textit{Empathy}, \textit{Relevance}, and \textit{Fluency}. Note that we evaluate each metric independently and the scores range between 1 and 5, in which 1 is "not at all" and 5 is "very much". 

We ask the human judges to evaluate each of the following categories from a 1 to 5 scale, where 5 is the best score.
\begin{itemize}
    \item Empathy / Sympathy: Did the responses from the LISTENER show understanding of the feelings of the SPEAKER talking about their experience?
    \item Relevance: Did the responses of the LISTENER seem appropriate to the conversation? Were they on-topic?  
    \item Fluency: Could you understand the responses from the LISTENER? Did the language seem accurate?
\end{itemize}

\paragraph{Human A/B Test} In this human evaluation task, we aim to directly compare the generated responses with each other. We randomly sample 100 dialogues each for \textit{MoEL vs \{TRS, Multi-TRS\}}. Three workers are given randomly ordered responses from either MoEL or \{TRS, Multi-TRS\}, and are prompted to choose the better response. They can either choose one of the responses or select \textit{tie} when the provided options are either both good or both bad.

\section{Results}
\paragraph{Emotion detection} To verify whether our model can attend to the appropriate listeners, we compute the emotion detection accuracy for each turn. Our model achieve $38\%$, $63\%$, $74\%$ in terms of top-1, top-3, top-5 detection accuracy over 32 emotions. We notice that some emotions frequently appear in similar context (e.g., Annoyed, Angry, Furious) which might degrade the detection accuracy. Figure~\ref{acc} shows the per class accuracy in the test set. We can see that by using top-5 the majority of the emotion achieve around 80\% accuracy. 

\paragraph{Response evaluation} Both automatic and human evaluation results are shown in Table ~\ref{tab:evaluation}. TRS achieves the highest BLEU score and \textit{Fluency} score but the lowest \textit{Empathy} and \textit{Relevance} score. This shows us that the responses generated by TRS are more generic but cannot accurately capture the user emotions. With the additional supervision on user emotions, multi-task training improves both \textit{Empathy} and \textit{Relevance} score, but it still degrades \textit{Fluency}. In contrast, MoEL achieves the highest \textit{Empathy} and \textit{Relevance} score. This suggests that the multi-expert strategy helps to capture the user emotional states and context simultaneously, and elicits a more appropriate response. The human A/B tests also confirm that the responses from our model are more preferred by human judges.

\section{Analysis}
In order to understand whether or how MoEL can effectively improve other baselines, learn each emotion, and properly react to them, we conduct three different analysis: model response comparison, listener analysis, and visualization of the emotion distribution $p$.

\paragraph{Model response comparison} The top part of Table~\ref{compare} compares the generated responses from MoEL and the two baselines on two different speaker emotional states. In the first example, MoEL captures the exact emotion of the speaker, by replying with "cleaning up dog poop is pretty \textbf{gross}", instead of "\textbf{horrible}" and "\textbf{scary}". In the second example, both TRS and Multi-TRS fail to understand that the speaker is disappointed about the failure of his interview, and they generate inappropriate responses. On the other hand, MoEL shows an empathetic response by comforting the speaker with "I am sure you will do great again". More examples can be find in the Appendix.

\paragraph{Listener analysis} To have a better understanding of how each listener learned to react to different context, we conduct a study of comparing responses produced by different listeners. To do so, we \textit{fix} the input dialogue context and we manually modify the attention vector distribution $p$ used to produce the response. We experiment with the correct listener and four other listeners: $\textbf{Listener}_{terrified}$, $\textbf{Listener}_{sad}$, $\textbf{Listener}_{excited}$, $\textbf{Listener}_{proud}$. Given the same context, we expect that different listeners will react differently, as this is our inductive bias. For example, $\textbf{Listener}_{sad}$ is optimized to comfort sad people, and $\textbf{Listener}_{\{excited, proud\}}$ share the positive emotions from the user. From the generation results in the bottom parts of Table~\ref{compare} we can see that the corresponding listeners can produce empathetic and relevant responses when they reasonably match the speaker emotions. However, when the expected emotion label is opposite to the selected listener, such as \textit{caring} and \textit{sad}, the response becomes emotionally inappropriate. 

Interestingly, in the last example, the \textit{sad} listener actually produces a more meaningful response by encouraging the speaker. This is due to the first part of the context which conveys a sad emotion. On the other hand, for the same example, the \textit{excited} listener responds with very relevant yet unsympathetic response. In addition, as many dialogue contexts contain multiple emotions, being able to capture them would lead to a better understanding of the speaker emotional state.


\paragraph{Visualization of Emotion Distribution} Finally, to understand how MoEL chooses the listener according to the context, we visualize the emotion distribution $p$ in Figure~\ref{viz}. In most of the cases, the model attends to the proper listeners (emotions), and generate a proper responses. This is confirmed also by the accuracy results shown in Figure~\ref{acc}. However, our model is sometimes focuses on parts of the dialogue context. For example, in the fifth example in Figure~\ref{viz}, the model fails to detect the real emotion of speaker as the context contains ``I was pretty \textbf{surprised}'' in its last turn. 

On the other hand, the last three rows of the heatmap indicate that the model learns to leverage \textbf{multiple} listeners to produce an empathetic response. For example, when the speaker talks about some criminals that shot one of his neighbors, MoEL successfully detects both \textit{annoyed} and \textit{afraid} emotions from the context, and replies with an appropriate response "that is horrible! i am glad you are okay!" that addresses both emotions. 
However, in the third row, the model produces "you" instead of "he" by mistake. Although the model is able to capture relevant emotions for this case, other emotions also have non-negligible weights which results in a smooth emotion distribution $p$ that confuses the meta listener from accurately generating a response.

\section{Conclusion \& Future Work}
In this paper, we propose a novel way to generate empathetic dialogue responses by using Mixture of Empathetic Listeners (MoEL). Differently from previous works, our model understand the user feelings and responds accordingly by learning specific listeners for each emotion. We benchmark our model in \textit{empathetic-dialogues} dataset~\cite{rashkin2018know}, which is a multi-turn open-domain conversation corpus grounded on emotional situations. Our experimental results show that MoEL is able to achieve competitive performance in the task with the advantage of being more interpretable than other conventional models. Finally, we show that our model is able to automatically select the correct emotional decoder and effectively generate an empathetic response.

One of the possible extensions of this work would be incorporating it with Persona~\cite{zhang2018personalizing} and task-oriented dialogue systems~\cite{,gao2018neural,madotto2018mem2seq,wu2019global, wu2017end,wu2018end,reddy2018multi, raghu-etal-2019-disentangling}. Having a persona would allow the system to have more consistent and personalized responses, and combining open-domain conversations with task-oriented dialogue systems would equip the system with more engaging conversational capabilities, hence resulting in a more versatile dialogue system.



\bibliography{emnlp-ijcnlp-2019}

\begin{thebibliography}{65}
\expandafter\ifx\csname natexlab\endcsname\relax\def\natexlab#1{#1}\fi

\bibitem[{Aljundi et~al.(2017)Aljundi, Chakravarty, and
  Tuytelaars}]{aljundi2017expert}
Rahaf Aljundi, Punarjay Chakravarty, and Tinne Tuytelaars. 2017.
\newblock Expert gate: Lifelong learning with a network of experts.
\newblock In \emph{Proceedings of the IEEE Conference on Computer Vision and
  Pattern Recognition}, pages 3366--3375.

\bibitem[{Bertero et~al.(2016)Bertero, Siddique, Wu, Wan, Chan, and
  Fung}]{bertero2016real}
Dario Bertero, Farhad~Bin Siddique, Chien-Sheng Wu, Yan Wan, Ricky Ho~Yin Chan,
  and Pascale Fung. 2016.
\newblock Real-time speech emotion and sentiment recognition for interactive
  dialogue systems.
\newblock In \emph{Proceedings of the 2016 Conference on Empirical Methods in
  Natural Language Processing}, pages 1042--1047.

\bibitem[{Cai et~al.(2018)Cai, Wang, Bi, Tu, Liu, Lam, and
  Shi}]{cai2018skeleton}
Deng Cai, Yan Wang, Victoria Bi, Zhaopeng Tu, Xiaojiang Liu, Wai Lam, and
  Shuming Shi. 2018.
\newblock Skeleton-to-response: Dialogue generation guided by retrieval memory.
\newblock \emph{arXiv preprint arXiv:1809.05296}.

\bibitem[{Chatterjee et~al.(2019{\natexlab{a}})Chatterjee, Gupta, Chinnakotla,
  Srikanth, Galley, and Agrawal}]{chatterjee2019understanding}
Ankush Chatterjee, Umang Gupta, Manoj~Kumar Chinnakotla, Radhakrishnan
  Srikanth, Michel Galley, and Puneet Agrawal. 2019{\natexlab{a}}.
\newblock Understanding emotions in text using deep learning and big data.
\newblock \emph{Computers in Human Behavior}, 93:309--317.

\bibitem[{Chatterjee et~al.(2019{\natexlab{b}})Chatterjee, Narahari, Joshi, and
  Agrawal}]{SemEval2019Task3}
Ankush Chatterjee, Kedhar~Nath Narahari, Meghana Joshi, and Puneet Agrawal.
  2019{\natexlab{b}}.
\newblock Semeval-2019 task 3: Emocontext: Contextual emotion detection in
  text.
\newblock In \emph{Proceedings of The 13th International Workshop on Semantic
  Evaluation (SemEval-2019)}, Minneapolis, Minnesota.

\bibitem[{Collobert et~al.(2002)Collobert, Bengio, and
  Bengio}]{collobert2002parallel}
Ronan Collobert, Samy Bengio, and Yoshua Bengio. 2002.
\newblock A parallel mixture of svms for very large scale problems.
\newblock In \emph{Advances in Neural Information Processing Systems}, pages
  633--640.

\bibitem[{Deisenroth and Ng(2015)}]{deisenroth2015distributed}
Marc Deisenroth and Jun~Wei Ng. 2015.
\newblock Distributed gaussian processes.
\newblock In \emph{International Conference on Machine Learning}, pages
  1481--1490.

\bibitem[{Devlin et~al.(2018)Devlin, Chang, Lee, and
  Toutanova}]{devlin2018bert}
Jacob Devlin, Ming-Wei Chang, Kenton Lee, and Kristina Toutanova. 2018.
\newblock Bert: Pre-training of deep bidirectional transformers for language
  understanding.
\newblock \emph{arXiv preprint arXiv:1810.04805}.

\bibitem[{Dinan et~al.(2019)Dinan, Logacheva, Malykh, Miller, Shuster, Urbanek,
  Kiela, Szlam, Serban, Lowe et~al.}]{dinan2019second}
Emily Dinan, Varvara Logacheva, Valentin Malykh, Alexander Miller, Kurt
  Shuster, Jack Urbanek, Douwe Kiela, Arthur Szlam, Iulian Serban, Ryan Lowe,
  et~al. 2019.
\newblock The second conversational intelligence challenge (convai2).
\newblock \emph{arXiv preprint arXiv:1902.00098}.

\bibitem[{Dinan et~al.(2018)Dinan, Roller, Shuster, Fan, Auli, and
  Weston}]{dinan2018wizard}
Emily Dinan, Stephen Roller, Kurt Shuster, Angela Fan, Michael Auli, and Jason
  Weston. 2018.
\newblock Wizard of wikipedia: Knowledge-powered conversational agents.
\newblock \emph{ICLR}.

\bibitem[{Fan et~al.(2018{\natexlab{a}})Fan, Lam, and Li}]{fan2018multi}
Yingruo Fan, Jacqueline~CK Lam, and Victor~OK Li. 2018{\natexlab{a}}.
\newblock Multi-region ensemble convolutional neural network for facial
  expression recognition.
\newblock In \emph{International Conference on Artificial Neural Networks},
  pages 84--94. Springer.

\bibitem[{Fan et~al.(2018{\natexlab{b}})Fan, Lam, and Li}]{fan2018unsupervised}
Yingruo Fan, Jacqueline~CK Lam, and Victor~OK Li. 2018{\natexlab{b}}.
\newblock Unsupervised domain adaptation with generative adversarial networks
  for facial emotion recognition.
\newblock In \emph{2018 IEEE International Conference on Big Data (Big Data)},
  pages 4460--4464. IEEE.

\bibitem[{Fan et~al.(2018{\natexlab{c}})Fan, Lam, and Li}]{fan2018video}
Yingruo Fan, Jacqueline~CK Lam, and Victor~OK Li. 2018{\natexlab{c}}.
\newblock Video-based emotion recognition using deeply-supervised neural
  networks.
\newblock In \emph{Proceedings of the 2018 on International Conference on
  Multimodal Interaction}, pages 584--588. ACM.

\bibitem[{Gao et~al.(2018)Gao, Galley, and Li}]{gao2018neural}
Jianfeng Gao, Michel Galley, and Lihong Li. 2018.
\newblock Neural approaches to conversational ai.
\newblock In \emph{The 41st International ACM SIGIR Conference on Research \&
  Development in Information Retrieval}, pages 1371--1374. ACM.

\bibitem[{Hancock et~al.(2019)Hancock, Bordes, Mazare, and
  Weston}]{hancock2019learning}
Braden Hancock, Antoine Bordes, Pierre-Emmanuel Mazare, and Jason Weston. 2019.
\newblock Learning from dialogue after deployment: Feed yourself, chatbot!
\newblock \emph{arXiv preprint arXiv:1901.05415}.

\bibitem[{Hu et~al.(2017)Hu, Yang, Liang, Salakhutdinov, and
  Xing}]{hu2017toward}
Zhiting Hu, Zichao Yang, Xiaodan Liang, Ruslan Salakhutdinov, and Eric~P Xing.
  2017.
\newblock Toward controlled generation of text.
\newblock In \emph{International Conference on Machine Learning}, pages
  1587--1596.

\bibitem[{Jacobs et~al.(1991)Jacobs, Jordan, Nowlan, Hinton
  et~al.}]{jacobs1991adaptive}
Robert~A Jacobs, Michael~I Jordan, Steven~J Nowlan, Geoffrey~E Hinton, et~al.
  1991.
\newblock Adaptive mixtures of local experts.
\newblock \emph{Neural computation}, 3(1):79--87.

\bibitem[{Jordan and Jacobs(1994)}]{jordan1994hierarchical}
Michael~I Jordan and Robert~A Jacobs. 1994.
\newblock Hierarchical mixtures of experts and the em algorithm.
\newblock \emph{Neural computation}, 6(2):181--214.

\bibitem[{Joshi et~al.(2017)Joshi, Mi, and Faltings}]{joshi2017personalization}
Chaitanya~K Joshi, Fei Mi, and Boi Faltings. 2017.
\newblock Personalization in goal-oriented dialog.
\newblock \emph{arXiv preprint arXiv:1706.07503}.

\bibitem[{Kaiser et~al.(2017)Kaiser, Gomez, Shazeer, Vaswani, Parmar, Jones,
  and Uszkoreit}]{kaiser2017one}
Lukasz Kaiser, Aidan~N Gomez, Noam Shazeer, Ashish Vaswani, Niki Parmar, Llion
  Jones, and Jakob Uszkoreit. 2017.
\newblock One model to learn them all.
\newblock \emph{arXiv preprint arXiv:1706.05137}.

\bibitem[{Kingma and Ba(2014)}]{kingma2014adam}
Diederik~P Kingma and Jimmy Ba. 2014.
\newblock Adam: A method for stochastic optimization.
\newblock \emph{arXiv preprint arXiv:1412.6980}.

\bibitem[{Kulikov et~al.(2018)Kulikov, Miller, Cho, and
  Weston}]{kulikov2018importance}
Ilya Kulikov, Alexander~H Miller, Kyunghyun Cho, and Jason Weston. 2018.
\newblock Importance of a search strategy in neural dialogue modelling.
\newblock \emph{arXiv preprint arXiv:1811.00907}.

\bibitem[{Lee et~al.(2019)Lee, Liu, and Fung}]{lee2019team}
Nayeon Lee, Zihan Liu, and Pascale Fung. 2019.
\newblock Team yeon-zi at semeval-2019 task 4: Hyperpartisan news detection by
  de-noising weakly-labeled data.
\newblock In \emph{Proceedings of the 13th International Workshop on Semantic
  Evaluation}, pages 1052--1056.

\bibitem[{Li et~al.(2016{\natexlab{a}})Li, Galley, Brockett, Gao, and
  Dolan}]{li2016diversity}
Jiwei Li, Michel Galley, Chris Brockett, Jianfeng Gao, and Bill Dolan.
  2016{\natexlab{a}}.
\newblock A diversity-promoting objective function for neural conversation
  models.
\newblock In \emph{Proceedings of the 2016 Conference of the North American
  Chapter of the Association for Computational Linguistics: Human Language
  Technologies}, pages 110--119.

\bibitem[{Li et~al.(2016{\natexlab{b}})Li, Galley, Brockett, Spithourakis, Gao,
  and Dolan}]{li2016persona}
Jiwei Li, Michel Galley, Chris Brockett, Georgios Spithourakis, Jianfeng Gao,
  and Bill Dolan. 2016{\natexlab{b}}.
\newblock A persona-based neural conversation model.
\newblock In \emph{Proceedings of the 54th Annual Meeting of the Association
  for Computational Linguistics (Volume 1: Long Papers)}, volume~1, pages
  994--1003.

\bibitem[{Li et~al.(2016{\natexlab{c}})Li, Monroe, Ritter, Jurafsky, Galley,
  and Gao}]{li2016deep}
Jiwei Li, Will Monroe, Alan Ritter, Dan Jurafsky, Michel Galley, and Jianfeng
  Gao. 2016{\natexlab{c}}.
\newblock Deep reinforcement learning for dialogue generation.
\newblock In \emph{Proceedings of the 2016 Conference on Empirical Methods in
  Natural Language Processing}, pages 1192--1202.

\bibitem[{Liu et~al.(2016)Liu, Lowe, Serban, Noseworthy, Charlin, and
  Pineau}]{liu2016not}
Chia-Wei Liu, Ryan Lowe, Iulian Serban, Mike Noseworthy, Laurent Charlin, and
  Joelle Pineau. 2016.
\newblock How not to evaluate your dialogue system: An empirical study of
  unsupervised evaluation metrics for dialogue response generation.
\newblock In \emph{Proceedings of the 2016 Conference on Empirical Methods in
  Natural Language Processing}, pages 2122--2132.

\bibitem[{Lubis et~al.(2018)Lubis, Sakti, Yoshino, and
  Nakamura}]{lubis2018eliciting}
Nurul Lubis, Sakriani Sakti, Koichiro Yoshino, and Satoshi Nakamura. 2018.
\newblock Eliciting positive emotion through affect-sensitive dialogue response
  generation: A neural network approach.
\newblock In \emph{Thirty-Second AAAI Conference on Artificial Intelligence}.

\bibitem[{Madotto et~al.(2019)Madotto, Lin, Wu, and
  Fung}]{madotto2019personalizing}
Andrea Madotto, Zhaojiang Lin, Chien-Sheng Wu, and Pascale Fung. 2019.
\newblock Personalizing dialogue agents via meta-learning.
\newblock In \emph{Proceedings of the 57th Conference of the Association for
  Computational Linguistics}, pages 5454--5459.

\bibitem[{Madotto et~al.(2018)Madotto, Wu, and Fung}]{madotto2018mem2seq}
Andrea Madotto, Chien-Sheng Wu, and Pascale Fung. 2018.
\newblock Mem2seq: Effectively incorporating knowledge bases into end-to-end
  task-oriented dialog systems.
\newblock In \emph{Proceedings of the 56th Annual Meeting of the Association
  for Computational Linguistics (Volume 1: Long Papers)}, pages 1468--1478.

\bibitem[{Mazare et~al.(2018{\natexlab{a}})Mazare, Humeau, Raison, and
  Bordes}]{mazare2018training}
Pierre-Emmanuel Mazare, Samuel Humeau, Martin Raison, and Antoine Bordes.
  2018{\natexlab{a}}.
\newblock Training millions of personalized dialogue agents.
\newblock In \emph{Proceedings of the 2018 Conference on Empirical Methods in
  Natural Language Processing}, pages 2775--2779.

\bibitem[{Mazare et~al.(2018{\natexlab{b}})Mazare, Humeau, Raison, and
  Bordes}]{millionspersona}
Pierre-Emmanuel Mazare, Samuel Humeau, Martin Raison, and Antoine Bordes.
  2018{\natexlab{b}}.
\newblock \href {http://aclweb.org/anthology/D18-1298} {Training millions of
  personalized dialogue agents}.
\newblock In \emph{Proceedings of the 2018 Conference on Empirical Methods in
  Natural Language Processing}, pages 2775--2779. Association for Computational
  Linguistics.

\bibitem[{Miller et~al.(2016)Miller, Fisch, Dodge, Karimi, Bordes, and
  Weston}]{miller2016key}
Alexander Miller, Adam Fisch, Jesse Dodge, Amir-Hossein Karimi, Antoine Bordes,
  and Jason Weston. 2016.
\newblock Key-value memory networks for directly reading documents.
\newblock In \emph{Proceedings of the 2016 Conference on Empirical Methods in
  Natural Language Processing}, pages 1400--1409.

\bibitem[{Papineni et~al.(2002)Papineni, Roukos, Ward, and
  Zhu}]{papineni2002bleu}
Kishore Papineni, Salim Roukos, Todd Ward, and Wei-Jing Zhu. 2002.
\newblock Bleu: a method for automatic evaluation of machine translation.
\newblock In \emph{Proceedings of the 40th annual meeting on association for
  computational linguistics}, pages 311--318. Association for Computational
  Linguistics.

\bibitem[{Pennington et~al.(2014)Pennington, Socher, and
  Manning}]{pennington2014glove}
Jeffrey Pennington, Richard Socher, and Christopher Manning. 2014.
\newblock Glove: Global vectors for word representation.
\newblock In \emph{Proceedings of the 2014 conference on empirical methods in
  natural language processing (EMNLP)}, pages 1532--1543.

\bibitem[{Raghu et~al.(2019)Raghu, Gupta, and
  {Mausam}}]{raghu-etal-2019-disentangling}
Dinesh Raghu, Nikhil Gupta, and {Mausam}. 2019.
\newblock \href {https://doi.org/10.18653/v1/N19-1126} {{D}isentangling
  {L}anguage and {K}nowledge in {T}ask-{O}riented {D}ialogs}.
\newblock In \emph{Proceedings of the 2019 Conference of the North {A}merican
  Chapter of the Association for Computational Linguistics: Human Language
  Technologies, Volume 1 (Long and Short Papers)}, pages 1239--1255,
  Minneapolis, Minnesota. Association for Computational Linguistics.

\bibitem[{Rashkin et~al.(2018)Rashkin, Smith, Li, and
  Boureau}]{rashkin2018know}
Hannah Rashkin, Eric~Michael Smith, Margaret Li, and Y-Lan Boureau. 2018.
\newblock I know the feeling: Learning to converse with empathy.
\newblock \emph{arXiv preprint arXiv:1811.00207}.

\bibitem[{Rasmussen and Ghahramani(2002)}]{rasmussen2002infinite}
Carl~E Rasmussen and Zoubin Ghahramani. 2002.
\newblock Infinite mixtures of gaussian process experts.
\newblock In \emph{Advances in neural information processing systems}, pages
  881--888.

\bibitem[{Reddy et~al.(2018)Reddy, Contractor, Raghu, and
  Joshi}]{reddy2018multi}
Revanth Reddy, Danish Contractor, Dinesh Raghu, and Sachindra Joshi. 2018.
\newblock Multi-level memory for task oriented dialogs.
\newblock \emph{arXiv preprint arXiv:1810.10647}.

\bibitem[{Schmidhuber(1987)}]{schmidhuber:1987:srl}
Jurgen Schmidhuber. 1987.
\newblock \href {http://www.idsia.ch/~juergen/diploma.html} {Evolutionary
  principles in self-referential learning. on learning now to learn: The
  meta-meta-meta...-hook}.
\newblock Diploma thesis, Technische Universitat Munchen, Germany, 14 May.

\bibitem[{Serban et~al.(2016)Serban, Lowe, Charlin, and
  Pineau}]{serban2016generative}
Iulian~Vlad Serban, Ryan Lowe, Laurent Charlin, and Joelle Pineau. 2016.
\newblock Generative deep neural networks for dialogue: A short review.
\newblock \emph{arXiv preprint arXiv:1611.06216}.

\bibitem[{Shahbaba and Neal(2009)}]{shahbaba2009nonlinear}
Babak Shahbaba and Radford Neal. 2009.
\newblock Nonlinear models using dirichlet process mixtures.
\newblock \emph{Journal of Machine Learning Research}, 10(Aug):1829--1850.

\bibitem[{Shazeer et~al.(2017)Shazeer, Mirhoseini, Maziarz, Davis, Le, Hinton,
  and Dean}]{shazeer2017outrageously}
Noam Shazeer, Azalia Mirhoseini, Krzysztof Maziarz, Andy Davis, Quoc Le,
  Geoffrey Hinton, and Jeff Dean. 2017.
\newblock Outrageously large neural networks: The sparsely-gated
  mixture-of-experts layer.
\newblock \emph{arXiv preprint arXiv:1701.06538}.

\bibitem[{Shin et~al.(2019)Shin, Xu, Madotto, and Fung}]{shin2019happybot}
Jamin Shin, Peng Xu, Andrea Madotto, and Pascale Fung. 2019.
\newblock Happybot: Generating empathetic dialogue responses by improving user
  experience look-ahead.
\newblock \emph{arXiv preprint arXiv:1906.08487}.

\bibitem[{Theis and Bethge(2015)}]{theis2015generative}
Lucas Theis and Matthias Bethge. 2015.
\newblock Generative image modeling using spatial lstms.
\newblock In \emph{Advances in Neural Information Processing Systems}, pages
  1927--1935.

\bibitem[{Tresp(2001)}]{tresp2001mixtures}
Volker Tresp. 2001.
\newblock Mixtures of gaussian processes.
\newblock In \emph{Advances in neural information processing systems}, pages
  654--660.

\bibitem[{Vaswani et~al.(2017)Vaswani, Shazeer, Parmar, Uszkoreit, Jones,
  Gomez, Kaiser, and Polosukhin}]{vaswani2017attention}
Ashish Vaswani, Noam Shazeer, Niki Parmar, Jakob Uszkoreit, Llion Jones,
  Aidan~N Gomez, {\L}ukasz Kaiser, and Illia Polosukhin. 2017.
\newblock Attention is all you need.
\newblock In \emph{Advances in neural information processing systems}, pages
  5998--6008.

\bibitem[{Vinyals and Le(2015)}]{vinyals2015neural}
Oriol Vinyals and Quoc Le. 2015.
\newblock A neural conversational model.
\newblock In \emph{International Conference on Machine Learning}.

\bibitem[{Wang and Wan(2018)}]{wang2018sentigan}
Ke~Wang and Xiaojun Wan. 2018.
\newblock Sentigan: generating sentimental texts via mixture adversarial
  networks.
\newblock In \emph{Proceedings of the 27th International Joint Conference on
  Artificial Intelligence}, pages 4446--4452. AAAI Press.

\bibitem[{Weston et~al.(2018)Weston, Dinan, and Miller}]{weston2018retrieve}
Jason Weston, Emily Dinan, and Alexander Miller. 2018.
\newblock Retrieve and refine: Improved sequence generation models for
  dialogue.
\newblock In \emph{Proceedings of the 2018 EMNLP Workshop SCAI: The 2nd
  International Workshop on Search-Oriented Conversational AI}, pages 87--92.

\bibitem[{Winata et~al.(2017)Winata, Kampman, Yang, Dey, and
  Fung}]{winata2017nora}
Genta~Indra Winata, Onno Kampman, Yang Yang, Anik Dey, and Pascale Fung. 2017.
\newblock Nora the empathetic psychologist.
\newblock \emph{Proc. Interspeech 2017}, pages 3437--3438.

\bibitem[{Winata et~al.(2019)Winata, Madotto, Lin, Shin, Xu, Xu, and
  Fung}]{winata2019caire_hkust}
Genta~Indra Winata, Andrea Madotto, Zhaojiang Lin, Jamin Shin, Yan Xu, Peng Xu,
  and Pascale Fung. 2019.
\newblock Caire\_hkust at semeval-2019 task 3: Hierarchical attention for
  dialogue emotion classification.
\newblock In \emph{Proceedings of the 13th International Workshop on Semantic
  Evaluation}, pages 142--147.

\bibitem[{Wolf et~al.(2019)Wolf, Sanh, Chaumond, and
  Delangue}]{wolf2019transfertransfo}
Thomas Wolf, Victor Sanh, Julien Chaumond, and Clement Delangue. 2019.
\newblock Transfertransfo: A transfer learning approach for neural network
  based conversational agents.
\newblock \emph{arXiv preprint arXiv:1901.08149}.

\bibitem[{Wu et~al.(2017)Wu, Madotto, Winata, and Fung}]{wu2017end}
Chien-Sheng Wu, Andrea Madotto, Genta Winata, and Pascale Fung. 2017.
\newblock End-to-end recurrent entity network for entity-value independent
  goal-oriented dialog learning.
\newblock In \emph{Dialog System Technology Challenges Workshop, DSTC6}.

\bibitem[{Wu et~al.(2018{\natexlab{a}})Wu, Madotto, Winata, and
  Fung}]{wu2018end}
Chien-Sheng Wu, Andrea Madotto, Genta~Indra Winata, and Pascale Fung.
  2018{\natexlab{a}}.
\newblock End-to-end dynamic query memory network for entity-value independent
  task-oriented dialog.
\newblock In \emph{2018 IEEE International Conference on Acoustics, Speech and
  Signal Processing (ICASSP)}, pages 6154--6158. IEEE.

\bibitem[{Wu et~al.(2019)Wu, Socher, and Xiong}]{wu2019global}
Chien-Sheng Wu, Richard Socher, and Caiming Xiong. 2019.
\newblock Global-to-local memory pointer networks for task-oriented dialogue.
\newblock \emph{arXiv preprint arXiv:1901.04713}.

\bibitem[{Wu et~al.(2018{\natexlab{b}})Wu, Wei, Huang, Li, and
  Zhou}]{wu2018response}
Yu~Wu, Furu Wei, Shaohan Huang, Zhoujun Li, and Ming Zhou. 2018{\natexlab{b}}.
\newblock Response generation by context-aware prototype editing.
\newblock \emph{arXiv preprint arXiv:1806.07042}.

\bibitem[{Xu et~al.(2018)Xu, Madotto, Wu, Park, and Fung}]{xu2018emo2vec}
Peng Xu, Andrea Madotto, Chien-Sheng Wu, Ji~Ho Park, and Pascale Fung. 2018.
\newblock Emo2vec: Learning generalized emotion representation by multi-task
  training.
\newblock In \emph{Proceedings of the 9th Workshop on Computational Approaches
  to Subjectivity, Sentiment and Social Media Analysis}, pages 292--298.

\bibitem[{Yao et~al.(2009)Yao, Walther, Beck, and
  Fei-Fei}]{yao2009hierarchical}
Bangpeng Yao, Dirk Walther, Diane Beck, and Li~Fei-Fei. 2009.
\newblock Hierarchical mixture of classification experts uncovers interactions
  between brain regions.
\newblock In \emph{Advances in Neural Information Processing Systems}, pages
  2178--2186.

\bibitem[{Yavuz et~al.(2018)Yavuz, Rastogi, Chao, Hakkani-T{\"u}r, and
  AI}]{yavuz2018deepcopy}
Semih Yavuz, Abhinav Rastogi, Guanlin Chao, Dilek Hakkani-T{\"u}r, and
  Amazon~Alexa AI. 2018.
\newblock Deepcopy: Grounded response generation with hierarchical pointer
  networks.
\newblock \emph{ConvAI Workshop@NIPS}.

\bibitem[{Zemlyanskiy and Sha(2018)}]{zemlyanskiy2018aiming}
Yury Zemlyanskiy and Fei Sha. 2018.
\newblock Aiming to know you better perhaps makes me a more engaging dialogue
  partner.
\newblock \emph{CoNLL 2018}, page 551.

\bibitem[{Zhang et~al.(2018{\natexlab{a}})Zhang, Dinan, Urbanek, Szlam, Kiela,
  and Weston}]{zhang2018personalizing}
Saizheng Zhang, Emily Dinan, Jack Urbanek, Arthur Szlam, Douwe Kiela, and Jason
  Weston. 2018{\natexlab{a}}.
\newblock Personalizing dialogue agents: I have a dog, do you have pets too?
\newblock In \emph{Proceedings of the 56th Annual Meeting of the Association
  for Computational Linguistics (Volume 1: Long Papers)}, pages 2204--2213.

\bibitem[{Zhang et~al.(2018{\natexlab{b}})Zhang, Dinan, Urbanek, Szlam, Kiela,
  and Weston}]{personachat}
Saizheng Zhang, Emily Dinan, Jack Urbanek, Arthur Szlam, Douwe Kiela, and Jason
  Weston. 2018{\natexlab{b}}.
\newblock \href {http://aclweb.org/anthology/P18-1205} {Personalizing dialogue
  agents: I have a dog, do you have pets too?}
\newblock In \emph{Proceedings of the 56th Annual Meeting of the Association
  for Computational Linguistics (Volume 1: Long Papers)}, pages 2204--2213.
  Association for Computational Linguistics.

\bibitem[{Zhou et~al.(2018)Zhou, Huang, Zhang, Zhu, and
  Liu}]{zhou2018emotional}
Hao Zhou, Minlie Huang, Tianyang Zhang, Xiaoyan Zhu, and Bing Liu. 2018.
\newblock Emotional chatting machine: Emotional conversation generation with
  internal and external memory.
\newblock In \emph{Thirty-Second AAAI Conference on Artificial Intelligence}.

\bibitem[{Zhou and Wang(2018)}]{zhou2018mojitalk}
Xianda Zhou and William~Yang Wang. 2018.
\newblock Mojitalk: Generating emotional responses at scale.
\newblock In \emph{Proceedings of the 56th Annual Meeting of the Association
  for Computational Linguistics (Volume 1: Long Papers)}, pages 1128--1137.

\end{thebibliography}
\bibliographystyle{acl_natbib}
\clearpage
\onecolumn
\appendix
\section{Supplemental Material}
\begin{table}[h]

\begin{tabular}{|p{1.8cm}|p{13cm}|}
\hline
\textbf{Label}     & Caring                                                                                                                                                                                                                             \\ \hline
\textbf{Context}   & \begin{tabular}[c]{@{}l@{}}\textbf{Speaker}: my boyfriend is sick . he can not eat much so i made some soup for him.\\ \textbf{Listener}: you are sweet , what kind of soup did you make him ? \\ \textbf{Speaker}: i made chicken noodle soup .\end{tabular} \\ \hline
\textbf{TRS}       & i am sorry to hear that . i hope he gets better !                                                                                                                                                                                  \\ \hline
\textbf{Multi-TRS} & that is good , i am sure he will do great !                                                                                                                                                                                        \\ \hline
\textbf{MoEL}      & well that is very nice of you . i hope he gets better !                                                                                                                                                                            \\ \hline
\textbf{Goal}      & i love chicken noodle , he must be so grateful .                                                                                                                                                                                   \\ \hline
\end{tabular}

\end{table}

\vspace{-25pt}
\begin{table}[h]
\begin{tabular}{|p{1.8cm}|p{13cm}|}
\hline
\textbf{Label}     & Nostalgic                                                                                                                                                                                                                            \\ \hline
\textbf{Context}   & \begin{tabular}[c]{@{}l@{}}\textbf{Speaker}: i do not watch a lot of kids movies these days , \\but they always bring back fond memories.\\ \textbf{Listener}: same . what are your favorites ? \\ \textbf{Speaker}: i have seen the movie trolls probably 100 times , that and harry potter !\end{tabular} \\ \hline
\textbf{TRS}       & i do not think i have ever had one of those .                                                                                                                                                                                 \\ \hline
\textbf{Multi-TRS} & i have never heard of them , but i am sure you will enjoy it .                                                                                                                                                                                    \\ \hline
\textbf{MoEL}      & i do too . i do that all the time .                                                                                                                                                                           \\ \hline
\textbf{Goal}      & that is funny . i have also seen trolls about 100 times . my daughter is now sick of it !                                                                                                                                                                                  \\ \hline
\end{tabular}
\begin{tabular}{|p{1.8cm}|p{13cm}|}

\hline
\textbf{Label}     & Grateful                                                              \\ \hline
\textbf{Context}   & \textbf{Speaker}: i just got a promotion at my job !                           \\ \hline
\textbf{TRS}       & that is great ! i hope you get it !                                   \\ \hline
\textbf{Multi-TRS} & that is great ! congratulations !,congratulations !,congratulations ! \\ \hline
\textbf{MoEL}      & that is great ! what kind of job is it ?                              \\ \hline
\textbf{Goal}      & i am happy for you . what kind of work                                \\ \hline
\end{tabular}
\end{table}

\vspace{-25pt}
\begin{table}[h]

\begin{tabular}{|p{1.8cm}|p{13cm}|}
\hline
\textbf{Label}     & Angry                                                                                                                                                                                                                        \\ \hline
\textbf{Context}   & \begin{tabular}[c]{@{}l@{}}\textbf{Speaker}: one of my coworkers really ticked me off one day . \\ she decided she would delegate some work to me by slamming on my desk and \\ demanding i do it . she had some nerve !\end{tabular} \\ \hline
\textbf{TRS}       & oh no ! i hate when that happens . i am sorry to hear that .                                                                                                                                                                 \\ \hline
\textbf{Multi-TRS} & oh no ! that is so scary ! i would be so worried about that .                                                                                                                                                                \\ \hline
\textbf{MoEL}      & that is awful . did you confront her ?                                                                                                                                                                                       \\ \hline
\textbf{Goal}      & did you get mad at her ?                                                                 \\ \hline
\end{tabular}
\caption{Generated responses from TRS, Multi-TRS and MoEL in different user emotion states. Goal responses are given for reference}
\end{table}




\end{document}